\documentclass[10pt,twocolumn,letterpaper]{article}

\usepackage[pagenumbers]{cvpr} 
\usepackage{multirow}

\definecolor{cvprblue}{rgb}{0.21,0.49,0.74}
\usepackage[pagebackref,breaklinks,colorlinks,allcolors=cvprblue]{hyperref}
\usepackage{pifont}
\usepackage[accsupp]{axessibility}

\def\paperID{10924} 
\def\confName{CVPR}
\def\confYear{2025}

\title{4DGC: Rate-Aware 4D Gaussian Compression for Efficient Streamable Free-Viewpoint Video}

\author{Qiang Hu$^{1}$$^{*}$
\and
Zihan Zheng$^{1}$$^{*}$
\and
Houqiang Zhong$^{2}$
\and
Sihua Fu$^1$
\and
Li Song$^{2}$
\and
XiaoyunZhang$^{1}$$^{\dagger}$ 
\qquad \qquad Guangtao Zhai$^{2}$ \qquad \qquad
Yanfeng Wang$^{3}$$^{\dagger}$   \\
$^{1}$ Cooperative Medianet Innovation Center, Shanghai Jiao Tong University \\ $^{2}$ School of Electronic Information and Electrical Engineering, Shanghai Jiao Tong University \\ $^{3}$ School of Artificial Intelligence, Shanghai Jiao Tong University \\
\vspace{-3em}
}

\begin{document}

\twocolumn[{%
\renewcommand\twocolumn[1][]{#1}%
\maketitle
\includegraphics[width=\linewidth]{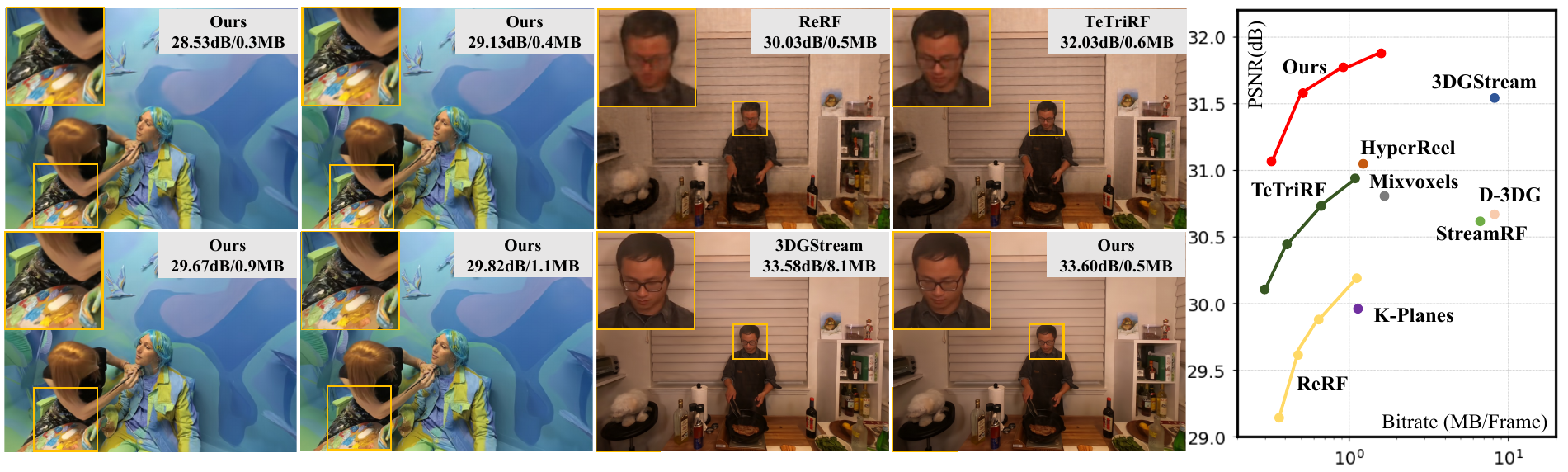}
\vspace{-2em}
\captionof{figure}{\textbf{Left}: 4DGC results, showcasing flexible quality levels across various bitrates. \textbf{Middle}: Comparison of visual quality and bitrate with state-of-the-art methods.  \textbf{Right}: The RD performance of our approach surpasses that of prior work (e.g. 3DGStream \cite{sun20243dgstream}, ReRF \cite{rerf}, TeTriRF \cite{tetrirf}). \vspace{1em}}
\label{fig:teaser}
}]

{\let\thefootnote\relax\footnote{\
        ${ }^{*}$ These authors contributed equally.
		${ }^{\dagger}$ The corresponding authors are Xiaoyun Zhang (xiaoyun.zhang@sjtu.edu.cn) and Yanfeng Wang (wangyanfeng622@sjtu.edu.cn).
}}\par

\vspace{-1em}
\begin{abstract}

3D Gaussian Splatting (3DGS) has substantial potential for enabling photorealistic Free-Viewpoint Video (FVV) experiences. However, the vast number of Gaussians and their associated attributes poses significant challenges for storage and transmission. Existing methods typically handle dynamic 3DGS representation and compression separately, neglecting motion information and the rate-distortion (RD) trade-off during training, leading to performance degradation and increased model redundancy.
To address this gap, we propose 4DGC, a novel rate-aware 4D Gaussian compression framework that  significantly reduces storage size while maintaining superior RD performance for FVV. Specifically, 4DGC introduces a motion-aware dynamic Gaussian representation that utilizes a compact motion grid combined with sparse compensated Gaussians to exploit inter-frame similarities. This representation effectively handles large motions, preserving quality and reducing temporal redundancy. Furthermore, we present an end-to-end compression scheme that employs differentiable quantization and a tiny implicit entropy model to compress the motion grid and compensated Gaussians efficiently. The entire framework is jointly optimized using a rate-distortion trade-off. Extensive experiments demonstrate that 4DGC supports variable bitrates and consistently outperforms existing methods in RD performance across multiple datasets.

\end{abstract}    

\vspace{-1em}
\section{Introduction}
\label{sec:intro}



Free-Viewpoint Video (FVV) enables immersive real-time navigation of scenes from any perspective, enhancing user engagement with high interactivity and realism. This makes FVV ideal for applications in entertainment, virtual reality, sports broadcasting, and telepresence. However, streaming and rendering high-quality FVV remains challenging, particularly for sequences with large motions, complex backgrounds, and extended durations. The primary difficulty lies in developing an efficient representation and compression method for FVV that supports streaming with limited bitrate while maintaining high fidelity.

Traditional approaches to FVV reconstruction have primarily relied on point cloud-based methods \cite{graziosi2020overview} and depth-based techniques \cite{boyce2021mpeg}, which struggle to deliver high rendering quality and realism, especially in complex scenes. Neural Radiance Fields (NeRF) and its variants \cite{nerf,DNeRF,streaming,tineuvox,HumanRF,kplanes,hexplane_2023_CVPR} have demonstrated impressive results in reconstructing FVV by learning continuous 3D scene representations, yet they face limitations in supporting long sequences and streaming. Recent approaches \cite{rerf, videorf, tetrirf, zheng2024jointrf, zheng2024hpc} address these issues by compressing explicit features of dynamic NeRF. For example, TeTriRF \cite{tetrirf} employs a hybrid representation with tri-planes to model dynamic scenes and applies a traditional video codec to further reduce redundancy. However,  these methods often suffer from slow training and rendering speeds.

Recently, 3D Gaussian Splatting (3DGS) \cite{kerbl3Dgaussians} has demonstrated exceptional rendering speed and quality compared to NeRF-based approaches for static scenes. Several methods \cite{Li_STG_2024_CVPR, Wu_2024_CVPR} have attempted to extend 3DGS to dynamic settings by incorporating temporal correspondence or time-dependency, but these approaches require loading all frames into memory for training and rendering, limiting their practicality for streaming applications. 3DGStream \cite{sun20243dgstream} models the inter-frame transformation and rotation of 3D Gaussians as a neural transformation cache, which reduces per-frame storage  requirements for FVV. However, the overall data volume remains substantial, hindering its ability to support efficient FVV transmission. Although a few studies \cite{hifi4g, jiang2024robust} have explored the compression of dynamic 3DGS, these methods heavily struggle with real-world dynamic scenes containing backgrounds, which limits their practical effectiveness. Additionally, they optimize representation and compression independently, overlooking the rate-distortion (RD) trade-off during training, which ultimately restricts compression efficiency.

In this paper, we propose 4DGC, a novel rate-aware compression method tailored for Gaussian-based FVV. Our key idea is to explicitly model motion between adjacent frames, estimate the bitrate of the 4D Gaussian representation during training, and incorporate rate and distortion terms into the loss function to enable end-to-end optimization. This allows us to achieve a compact and compression-friendly representation with optimal RD performance.  We realize this through two main innovations.  First, we introduce a motion-aware dynamic Gaussian representation for inter-frame modeling within long sequences. We train the 3DGS on the first frame (keyframe) to obtain the initial reference Gaussians. For each subsequent frame, 4DGC utilizes a compact multi-resolution motion grid to estimate the rigid motion of each Gaussian from the previous frame to the current one. Additionally, compensated Gaussians are sparsely added to account for newly observed regions or objects, further enhancing the accuracy of the representation. By leveraging inter-frame similarities, 4DGC effectively reduces temporal redundancy and ensures that the representation remains compact while maintaining high visual fidelity.

Second, we propose a unified end-to-end compression scheme that efficiently encodes the initial Gaussian spherical harmonics (SH) coefficients, the motion grid, and the compensated Gaussian SH coefficients. This compression approach incorporates differentiable quantization to facilitate gradient back-propagation and a tiny implicit entropy model for accurate bitrate estimation.  By optimizing the entire scheme through a rate-distortion trade-off during training, we significantly enhance compression performance. Experimental results show that our 4DGC supports variable bitrates and achieves state-of-the-art RD performance across various datasets. Compared to the SOTA method 3DGStream \cite{sun20243dgstream}, our approach achieves approximately a \textbf{16x} compression rate without quality degradation, as illustrated in Fig. \ref{fig:teaser}.

 In summary, our key contributions include:

\begin{itemize}
\item We present a compact and compression-friendly 4D Gaussian representation for streamable FVV that effectively captures dynamic motion and compensates for newly emerging objects, minimizing temporal redundancy and enhancing reconstruction quality.

\item We introduce an end-to-end 4D Gaussian compression framework that jointly optimizes representation and entropy models using a rate-distortion loss function, ensuring a low-entropy 4D Gaussian representation and significantly enhancing RD performance.


\item Extensive experimental results on the real-world datasets demonstrate that our 4DGC achieves superior reconstruction quality, bitrate efficiency, training time, and rendering speed compared to existing state-of-the-art dynamic scene compression methods. 

\end{itemize}

\section{Related Work}
\label{sec:relatedwork}

\begin{figure*}[t]
\centering
\includegraphics[width=\linewidth]{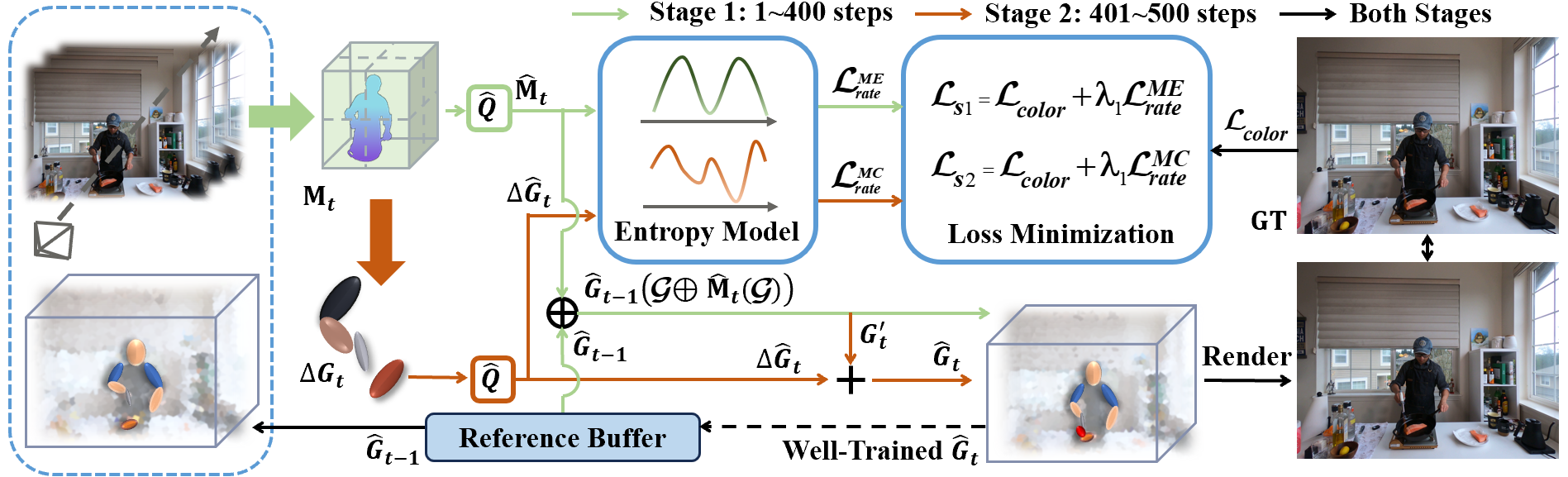}
\vspace{-1em}
\caption{Illustration of the 4DGC Framework. The reconstructed Gaussians from the previous frame, $\hat{\mathbf{G}}_{t-1}$, are retrieved from the reference buffer and combined with the input images of the current frame to facilitate learning of the motion grid $\mathbf{M}_t$ and the compensated Gaussians $\Delta \mathbf{G}_t$ through a two-stage training process. In the first stage, the motion grid and its associated entropy model are optimized. In the second stage, the compensated Gaussians are refined along with their corresponding entropy model. Both stages are supervised by a rate-distortion trade-off, employing simulated quantization and an entropy model to jointly optimize representation and compression.}
\label{fig:method} 
\vspace{-1.5em}
\end{figure*}

\textbf{Dynamic Modeling with NeRF.} Building on NeRF's success in static scene synthesis \cite{nerf,instant-ngp,merf,fpo++,DBARF_Chen_2023_CVPR,park2023camp,martinbrualla2020nerfw,barron2021mipnerf,barron2022mipnerf360,barron2023zipnerf}, several works have extended these methods to dynamic scenes.
Flow-based methods \cite{9578364,Li_2023_CVPR} construct 3D features from monocular videos with impressive results, at the cost of more extra priors like depth and motion for complex scenes.
Deformation field methods \cite{Nerfies,nerfplayer,DNeRF,NeuralRadianceFlow,streaming} warp frames to a canonical space to capture temporal features but suffer from slow training and rendering speeds. To accelerate the speeds,  some methods \cite{tineuvox,HumanRF,kplanes,hexplane_2023_CVPR,tensor4d,li2022neural,Park2023TemporalII,10377294,9879176} extend the radiance field into four dimensions using grid representation, plane-based representation or tensor factorization. However, these methods typically suffer from storage efficiency and are not suitable for streaming.

\textbf{Dynamic Modeling with 3DGS.} Recent advancements in 3DGS \cite{kerbl3Dgaussians} and its variants \cite{Huang2DGS2024, feng2024flashgsefficient3dgaussian, höllein20243dgslmfastergaussiansplattingoptimization, charatan23pixelsplat, gao20246dgsenhanceddirectionawaregaussian} have achieved photorealistic static scene rendering with high efficiency. However, for dynamic scenes, the per-frame 3DGS approach neglects temporal consistency, causing visual artifacts and model size growth. Some methods \cite{Li_STG_2024_CVPR,Wu_2024_CVPR,yang2023deformable3dgs,Guo2024Motionaware3G,yan20244d,yang2023gs4d} model Gaussian attributes over time to represent dynamic scenes as a unified model, improving quality but requiring the simultaneous loading of all data, which limits practical use in long-sequence streaming. Other methods \cite{luiten2023dynamic,sun20243dgstream,Guo2024Motionaware3G} track Gaussian motion frame by frame, which is suitable for streaming, but the large size of each frame hinders transmission efficiency. In contrast, our approach employs a compact multi-resolution motion grid combined with per-frame Gaussian compensation, reducing temporal redundancy and enhancing reconstruction quality.

\textbf{Dynamic Scene Compression.}  Recent advances in deep learning-based image and video compression \cite{balle2016end, balle2018variational, 8578437, 9204799, 8578795, 8493529, 8416591, 8708943, 8908826, 8953892, 9941493, 10003249, 8610323, agustsson2017soft} have demonstrated strong RD performance. In FVV compression, current approaches \cite{rerf,videorf,tetrirf,nerfplayer,peng2023representing,cut,miniwave,zheng2024jointrf,zheng2024hpc} primarily focus on compressing dynamic NeRF features to improve storage and transmission efficiency. Techniques like ReRF \cite{rerf} and TeTriRF \cite{tetrirf} apply traditional image/video encoding methods to dynamic scenes without end-to-end optimization, sacrificing dynamic detail and compression efficiency. Some approaches \cite{zheng2024jointrf, zheng2024hpc} achieve end-to-end optimization but struggle with scalability in open scenes and slow rendering. For 3DGS-based methods, most \cite{navaneet2023compact3d,fan2023lightgaussian,scaffoldgs,wang2024contextgs} focus on static scene compression, while dynamic scene techniques \cite{hifi4g,jiang2024robust} remain limited and typically support only background-free scenarios without comprehensive optimization. Our method achieves both high RD performance and fast decoding and rendering times in real-world scenarios thanks to our proposed motion-aware dynamic Gaussian representation and end-to-end joint compression.


\section{Method}

%

In this section, we introduce the details of the 4DGC framework. Fig. \ref{fig:method} illustrates the overall architecture of 4DGC. Our approach begins with a motion-aware dynamic Gaussian representation, composed of a compact motion grid and sparse compensated Gaussians (Sec. \ref{sec 3.1}). Subsequently, we describe a two-stage method combining motion estimation and Gaussian compensation to generate this representation, effectively capturing both spatial and temporal variations (Sec. \ref{sec 3.2}). Finally, we introduce an end-to-end compression scheme that jointly optimizes representation and entropy models, ensuring a low-entropy representation and greatly improving RD performance (Sec. \ref{sec 3.3}).

\subsection{Motion-aware Dynamic Gaussian Modeling}
\label{sec 3.1}

Recall that 3DGS represents scenes using a collection $\mathbf{G}$ of Gaussian primitives as an explicit representation similar to point clouds. Each Gaussian $\boldsymbol{\mathcal{G}} \in \mathbf{G}$ is defined by a set of optimizable parameters $\{ \boldsymbol{\mu}; \mathbf{R}; \mathbf{f}; \mathbf{s};\alpha \}$, where $\boldsymbol{\mu}$ is the center location, $\mathbf{R}$ is the rotation matrix, $\mathbf{f}$ represents SH coefficients for view-dependent color $\mathbf{c}$, $\mathbf{s}$ is the scaling vector, and $\alpha$ is the opacity. 
For a point $\mathbf{x}$ located within a Gaussian primitive, the spatial distribution is determined by $\boldsymbol{\mathcal{G}}(\mathbf{x})$,
\begin{align}
\boldsymbol{\mathcal{G}}(\mathbf{x}) = \exp\left(-\frac{1}{2}(\mathbf{x}- \boldsymbol{\mu})^T\mathbf{\Sigma}^{-1}(\mathbf{x}-\boldsymbol{\mu})\right)  \label{equ_3dg_distribution}  
\end{align}
where $\mathbf{\Sigma} = \mathbf{R}\mathbf{s}\mathbf{s}^T\mathbf{R}^T$. 
The rendering color $\mathbf{c}$ of a pixel is computed by alpha-blending the $N$ Gaussians overlapping the pixel in depth order:
\begin{align}
    \mathbf{c} = \sum_{i \in N} \mathbf{c}_i {\alpha'}_{i} \prod_{j=1}^{i-1} (1 - {\alpha'}_j)
\end{align}
where $\alpha_{i}'$ is the projection from the $i$-th Gaussian opacity onto the image plane, $\mathbf{c}_i$ is the $i$-th Gaussian color in viewing direction.



When extending the representation from static to dynamic scenes, a straightforward approach is stacking frame-wise static Gaussians to form a dynamic sequence. However, this method neglects temporal coherence, resulting in significant temporal redundancy, particularly in scenes with dense Gaussian primitives. Alternative methods \cite{Li_STG_2024_CVPR, Wu_2024_CVPR} extend Gaussians to 4D space for modeling the entire dynamic scene. Such approaches suffer from performance degradation in long sequences and are not suitable for streaming applications. To overcome these limitations, we propose a motion-aware dynamic Gaussian representation,  which explicitly models and tracks motion between adjacent frames to maintain spatial and temporal coherence.
\begin{figure}
\centering
\includegraphics[width=\linewidth]{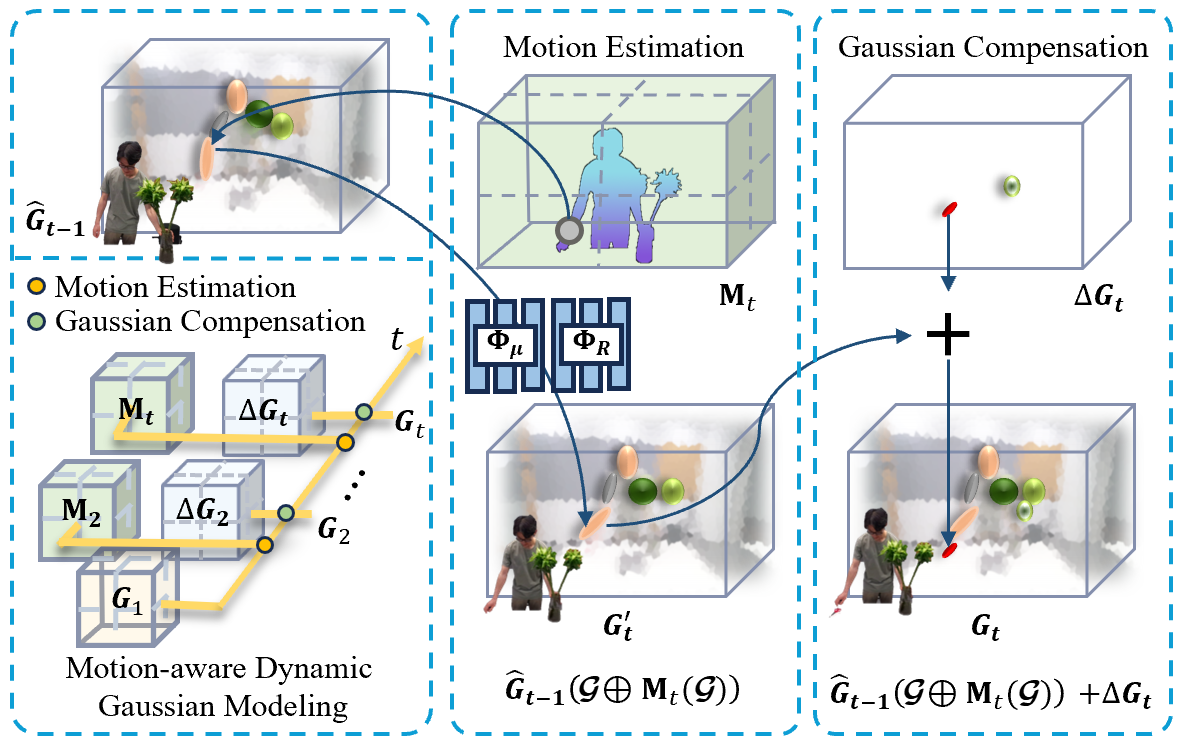}
\caption{Illustration of our motion-aware dynamic Gaussian modeling that utilizes a multi-resolution motion grid $\mathbf{M}_t$ with sparse compensated Gaussians $\Delta \mathbf{G}_t$ to exploit inter-frame similarities.}
\label{fig:motion}  
\vspace{-2em}
\end{figure}

Our modeling approach employs a complete 3DGS representation as the initial Gaussians $\mathbf{G}_1$ for the first frame (keyframe). For each subsequent frame, we utilize a multi-resolution motion grid $\mathbf{M}_t$ with two shared global lightweight MLPs, $\Phi_{\boldsymbol{\mu}}$ and $\Phi_{\mathbf{R}}$, to estimate the rigid motion of each Gaussian from the previous frame to the current one. This grid captures the multi-scale nature of motion, enabling precise modeling even for objects that move at varying speeds or directions. However, rigid transformation alone is insufficient for accurately representing newly emerging regions.  To address this, we dynamically add sparse compensated Gaussians $\Delta \mathbf{G}_t$ to account for newly observed regions in the current frame. Finally, our 4DGC sequentially represents the dynamic scene with N frames as $\mathbf{G}_1, \{\mathbf{M}_t, \Delta \mathbf{G}_t\}_{t=2}^{N}$, $\Phi_{\boldsymbol{\mu}}$, and $\Phi_{\mathbf{R}}$ as shown in Fig. \ref{fig:motion}. A major benefit of this design is that 4DGC fully exploits inter-frame similarities, reducing temporal redundancy and enhancing reconstruction quality.

\subsection{Sequential Representation Generation}
\label{sec 3.2}


Here, we introduce our two-stage scheme, which combines motion estimation and Gaussian compensation to generate a dynamic Gaussian representation that effectively captures both spatial and temporal variations in the scene. This process begins with motion estimation, which tracks and models frame-by-frame transformations in both translation and rotation, reducing inter-frame redundancy. To address scenarios where newly emerging objects or complex motion cannot be fully captured through estimation alone, a gaussian compensation step is applied to refine representation quality in suboptimal areas. Together, these stages form a flexible and high-fidelity representation that improves compression efficiency and supports high-quality rendering for streamable applications. We detail each stage below.

\textbf{Motion Estimation.}  For each inter-frame, we load the reconstructed Gaussians of the previous frame $\hat{\mathbf{G}}_{t-1}$ from the reference buffer, providing a stable reference for tracking transformations in the current frame. By combining these reference Gaussians with the input images of the current frame, we employ a motion grid, $\mathbf{M}_t$, along with two shared lightweight MLPs, $\Phi_{\boldsymbol{\mu}}$ and $\Phi_{\mathbf{R}}$, to predict the translation ($\Delta \boldsymbol{\mu}_t$) and rotation ($\Delta \mathbf{R}_t$) for each Gaussian. Specifically, to achieve accurate motion estimation, we use a multi-resolution motion grid $\mathbf{M}_t = \{\mathbf{M}_t^{l}\}_{l=1}^L$, where $L$ denotes the number of resolution levels,  to capture complex motions across various scales. For a Gaussian primitive $\boldsymbol{\mathcal{G}} \in \hat{\mathbf{G}}_{t-1}$  in the previous reconstructed frame, its center location $\boldsymbol{\mu}_{t-1}$ is mapped to multiple frequency bands $\mathbf{P}_{t-1}$ via positional encoding:
\begin{equation}
\begin{aligned}
    \mathbf{P}_{t-1} = \{\mathbf{P}_{t-1}^l\}_{l=1}^L = \{\sin(2^l\pi \boldsymbol{\mu}_{t-1}), \cos(2^l\pi \boldsymbol{\mu}_{t-1})\}_{l=1}^L
\end{aligned}    
\end{equation}
At each level, we use the mapped position to perform trilinear interpolation on $\mathbf{M}_t$
, producing motion features across different scales. 
These multi-scale features are then concatenated and fed into two lightweight MLPs, $\Phi_{\boldsymbol{\mu}}$ and $\Phi_{\mathbf{R}}$, to compute translation $\Delta \boldsymbol{\mu}_t$ and rotation $\Delta \mathbf{R}_t$ for each Gaussian. Thus, our motion estimation is formalized as follows:
\begin{equation}
\begin{aligned}
    \Delta \boldsymbol{\mu}_t &= \Phi_{\boldsymbol{\mu}}\left(\bigcup_{l=1}^L \text{interp}(\mathbf{P}_{t-1}^l, \mathbf{M}_{t}^l)\right) \\
   \Delta \mathbf{R}_t &= \Phi_{\mathbf{R}}\left(\bigcup_{l=1}^L \text{interp}(\mathbf{P}_{t-1}^l, \mathbf{M}_{t}^l)\right)
\end{aligned}
\end{equation}
where $\text{interp}(\cdot)$ represents the grid interpolation operation.

With the translation $\Delta \boldsymbol{\mu}_t$ and rotation $\Delta \mathbf{R}_t$, transformations are applied to each Gaussian in $\hat{\mathbf{G}}_{t-1}$, achieving smooth alignment from the previous to the current frame: 
\begin{equation}
\begin{aligned}
    \mathbf{G}_{t}^{'} &=  \hat{\mathbf{G}}_{t-1}(\boldsymbol{\mathcal{G}} \oplus \mathbf{M}_t(\boldsymbol{\mathcal{G}}))  \\
        &= \{ \boldsymbol{\mathcal{G}}(\boldsymbol{\mu}_{t-1}+ \Delta \boldsymbol{\mu}_t; \Delta \mathbf{R}_t \mathbf{R}_{t-1} ; C ) \mid \boldsymbol{\mathcal{G}} \in \hat{\mathbf{G}}_{t-1}\} 
\end{aligned} \label{eq:ME}
\end{equation}
where $\mathbf{G}_{t}^{'}$ denotes the transformed Gaussians for the current frame, $C$ represents the fixed parameters including $ \mathbf{f}_{t-1},\mathbf{s}_{t-1} $, and $ \alpha_{t-1}$. $\oplus$ denotes the operation of updating position $\boldsymbol{\mu}$ and rotation $\mathbf{R}$ for each Gaussian $\boldsymbol{\mathcal{G}}$ in $\hat{\mathbf{G}}_{t-1}$ according to ${\mathbf{M}}_t$. This hierarchical approach achieves precise motion prediction across multiple scales, capturing essential transformations and effectively reducing inter-frame redundancy.

\textbf{Gaussian Compensation.} Although motion estimation effectively captures the dynamics of previously observed objects within a scene, we found that relying solely on motion estimation is insufficient for achieving high-quality detail, particularly in cases involving newly emerging objects and subtle motion transformations. To address this limitation, we adopt a Gaussian compensation strategy, refining representation quality by integrating sparse compensated Gaussians $\Delta \mathbf{G}_t$ into $\mathbf{G}_{t}^{'}$ in suboptimal regions. 

We first identify the suboptimal areas requiring compensation. These regions are classified into two primary types: (1) regions with significant gradient changes, typically corresponding to newly appearing objects or scene edges, and (2) larger Gaussian primitives undergoing rapid transformations, leading to multiview perspective differences. For the first type, we apply gradient thresholding, cloning the Gaussian primitives at locations where the gradient exceeds a predefined threshold $\tau_g$ as $\Delta \mathbf{G}_t^g$, ensuring accurate representation of newly observed elements. 
For the second type, involving larger Gaussians impacted by rapid transformations, we clone two additional Gaussian primitives from the original Gaussian when its motion parameters exceed specified thresholds: $\tau_{\mu}$ for translation $|\Delta \boldsymbol{\mu}_t|$ and $\tau_R$ for rotation $|\Delta \mathbf{R}_t|$. The scale of two cloned Gaussians is reduced to $\frac{\mathbf{s}}{100}$ to capture detailed motion dynamics more precisely.

These newly compensated Gaussians are sampled in $\mathcal{N}(\boldsymbol{\mu}, 2 \boldsymbol{\Sigma}) $ around the original Gaussian and optimized in the second training stage. This Gaussian splitting process yields a fine-grained and adaptive representation, capturing complex motion patterns and enhancing continuity across frames. Overall, this compensation strategy significantly improves detail accuracy, reduces artifacts, and ensures high-quality reconstruction of dynamic scenes.

\subsection{End-to-end Joint Compression}
\label{sec 3.3}

We also propose an end-to-end 4D Gaussian compression framework that jointly optimizes representation and entropy models through a two-stage training process. To enable gradient back-propagation, we employ differentiable quantization, along with a compact implicit entropy model for accurate bitrate estimation of the motion grid $\mathbf{M}_t$ and compensated Gaussians $\Delta \mathbf{G}_t$. The first stage focuses on optimizing the motion grid alongside its associated entropy model, while the second stage refines the compensated Gaussians with their corresponding entropy model. Each stage is guided by a rate-distortion trade-off, ensuring a low-entropy 4D Gaussian representation and substantially improving RD performance.

\textbf{Simulated Quantization \& Rate Estimation.} Quantization and entropy coding effectively reduce the bitrate during compression at the expense of some information loss. However, the rounding operation in quantization prevents gradient propagation, which is incompatible with end-to-end training. To address this, we implement a differentiable quantization strategy using simulated quantization noise. Specifically, uniform noise $u \sim U\left(-\frac{1}{2q}, \frac{1}{2q}\right)$  is added to simulate quantization effects with step size $q$, enabling robust training while preserving gradient flow. For rate estimation, we use a tiny and trainable implicit entropy model \cite{begaint2020compressai} to approximate the probability mass function (PMF) of the quantized values, $\hat{y}$. Unlike the learned entropy model in image compression \cite{minnen2018joint}, which is learned from large-scale training datasets, our implicit entropy model is learned on-the-fly with the corresponding 4D Gaussian representation in the training.  The PMF is derived using the cumulative distribution function (CDF) as follows:
\begin{equation}
P_{PMF}(\hat{y}) = P_{CDF}(\hat{y} + \frac{1}{2}) - P_{CDF}(\hat{y} - \frac{1}{2}). \label{PMF}
\end{equation}
Incorporating this rate estimation into the loss function enables the network to learn feature distributions with inherently lower entropy. This effectively imposes a bitrate constraint during training while ensuring the compatibility of gradient-based optimization, balancing compression efficiency and model accuracy. 

\begin{figure*}[t]
\centering
\includegraphics[width=\linewidth]{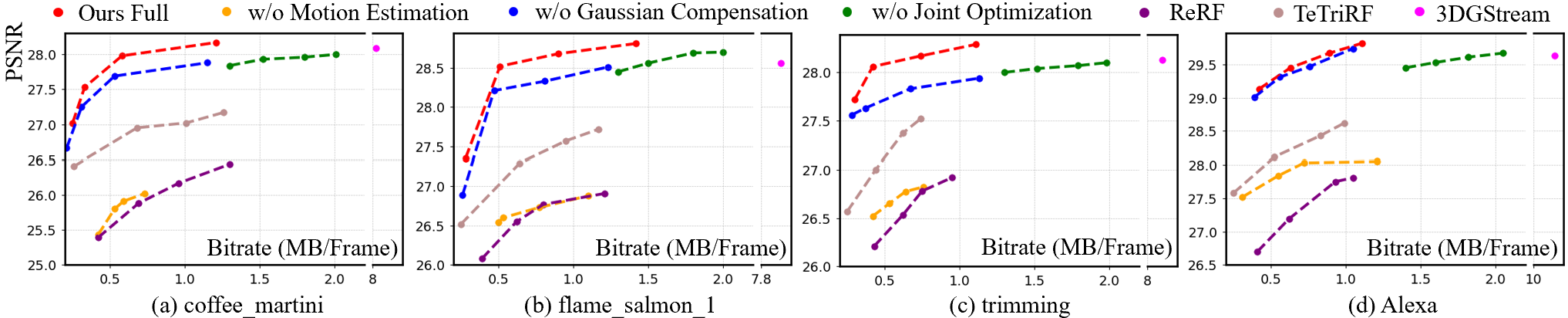}
\vspace{-2em}
\caption{Rate-distortion curves across different datasets. Rate-distortion curves not only illustrate the superiority of our method over ReRF \cite{rerf}, TeTriRF \cite{tetrirf}, and 3DGStream \cite{sun20243dgstream}, but also demonstrate the efficiency of various components within our method.}
\label{fig:ablation_rd}  
\vspace{-1.5em}
\end{figure*}

\textbf{Stage 1: Motion Grid Compression.}
In the first training stage, we jointly optimize  $\Phi_{\boldsymbol{\mu}}$, $\Phi_{\mathbf{R}}$, the multi-resolution motion grid $\mathbf{M}_t$ and its corresponding entropy model. This process enhances motion prediction accuracy while encouraging low-entropy characteristics in $\mathbf{M}_t$. 
Specifically, we apply simulated quantization to discretize the motion grid, ensuring compatibility with entropy encoding. The entropy model then assigns probabilities to each quantized element based on a learned probability mass function to estimate the bitrate of the motion grid more effectively. The loss function $\mathcal{L}_{s1}$ of this stage comprises a photometric term $\mathcal{L}_{color}$ and a rate term $\mathcal{L}_{rate}^{ME}$:
\begin{align}
    & \mathcal{L}_{s1} = \mathcal{L}_{color} + \lambda_1 \mathcal{L}_{rate}^{ME}\\
    & \mathcal{L}_{color} = (1 - \lambda_2) \|\mathbf{c}_g  - \hat{\mathbf{c}}  \|_1 + \lambda_2\mathcal{L}_{SSIM} \\
    & \mathcal{L}_{rate}^{ME} = -\frac{1}{N}\sum_{\hat{y} \in \hat{ \mathbf{M}}_t }{\log_2\left(P_{PMF}^1(\hat{y})\right)}
\end{align}
where $\mathcal{L}_{rate}^{ME}$ represents the estimated rate derived from $\hat{\mathbf{M}}_t$. $\mathbf{c}_g$ and $\hat{\mathbf{c}}$ refer to the ground truth and reconstructed colors , respectively. $ \mathcal{L}_{SSIM} $ is the D-SSIM\cite{dssim} metric between ground truth and the result rendered by 4DGC, and $ \lambda_2 $  serves as a weight parameter.  The parameter $\lambda_1$ balances the trade-off between rate and distortion, thus controlling model size and reconstruction quality.


\textbf{Stage 2: Compensated Gaussians Compression.} 
In the second training stage, we focus on optimizing the compensated Gaussians $\Delta \mathbf{G}_t$ and their entropy model to enhance detail capture and compression efficiency. While attributes like position and rotation are crucial for rendering, they require little storage. Thus, the main emphasis is on compressing the SH coefficients, which account for the largest storage cost.

Leveraging the trained motion grid $\mathbf{M}_t$ from Stage 1, we transform Gaussian primitives from the previous frame to the current frame while preserving fixed attributes like position, rotation, and scale. To enhance representation fidelity, we augment these transformed Gaussians with compensated Gaussians $\Delta \mathbf{G}_t$. The SH coefficients of $\Delta \mathbf{G}_t$ undergo simulated quantization and are processed by an implicit entropy model for accurate rate estimation. The loss function $\mathcal{L}_{s2}$ optimizes this compression process.
\begin{align}
    & \mathcal{L}_{s2} = \mathcal{L}_{color} + \lambda_1 \mathcal{L}_{rate}^{MC}\\
    & \mathcal{L}_{rate}^{MC} = -\frac{1}{M}\sum_{\hat{y} \in \hat{\mathbf{f}}_t^C }{\log_2\left(P_{PMF}^2(\hat{y})\right)}
\end{align}
where $\hat{\mathbf{f}}_t^C$ represents the quantized SH coefficients of the compensated Gaussians.This strategy is similarly applied to the initial Gaussians $\mathbf{G}_{1}$ in the keyframe. The joint optimization approach for representation and compression results in a compact and high-quality 4D Gaussian representation, facilitating efficient storage and transmission for FVV applications.

Once the training of the current frame is finished, we reconstruct the complete Gaussian representation for the current frame, $\hat{\mathbf{G}}_t$, as follows: 
\begin{align}
    \hat{\mathbf{G}}_t = \hat{\mathbf{G}}_{t-1}(\boldsymbol{\mathcal{G}} \oplus \hat{\mathbf{M}}_t(\boldsymbol{\mathcal{G}})) + \Delta \hat{\mathbf{G}}_t
\end{align}
where $\hat{\mathbf{M}}_t$ and $\Delta \hat{\mathbf{G}}_t$ represent the reconstructed motion grid and compensated Gaussians, respectively. 
Finally, $\hat{\mathbf{G}}_{t}$ is stored in the reference buffer to facilitate the reconstruction of the next frame. 


\begin{figure*}[t]
\centering
\includegraphics[width=\linewidth]{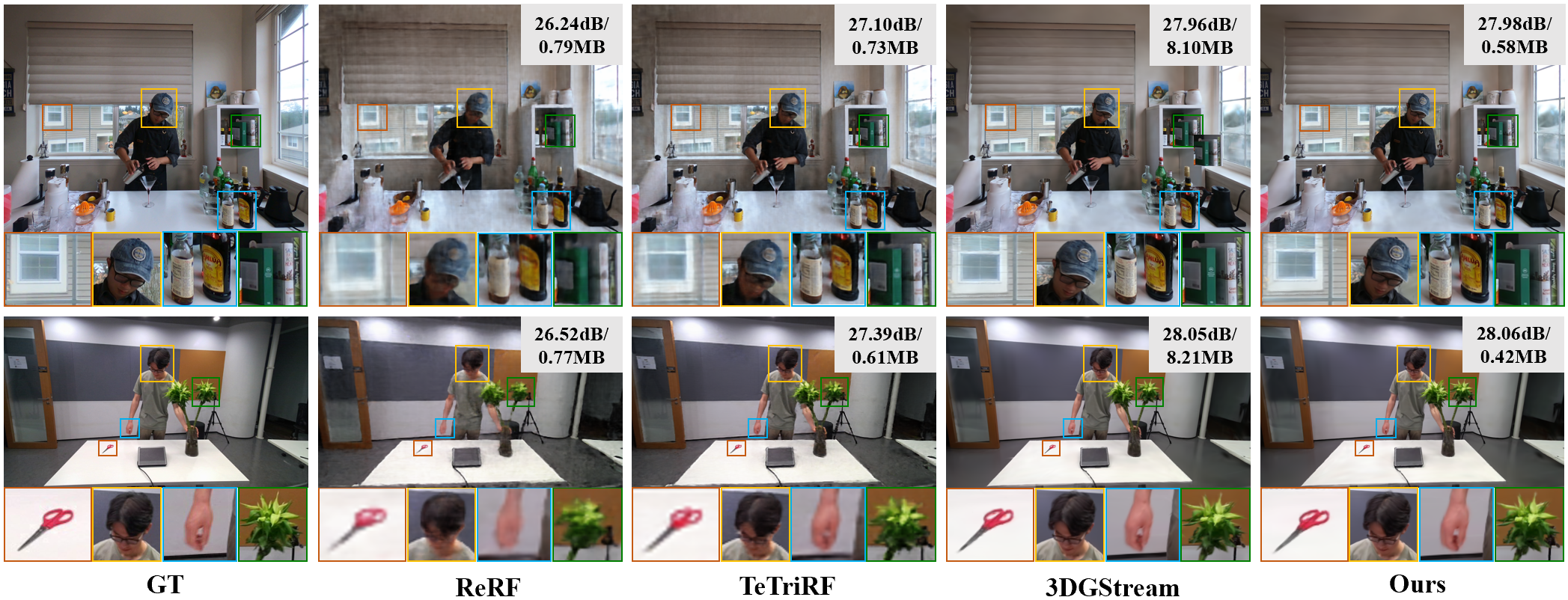}
\vspace{-2em}
\caption{Qualitative comparison on the N3DV \cite{li2022neural} and MeetRoom \cite{streaming} datasets against ReRF \cite{rerf}, TeTriRF \cite{tetrirf}, and 3DGStream \cite{sun20243dgstream}.}
\label{fig:Comparison}
\vspace{-1.5em}
\end{figure*}

\section{Experiments}

\subsection{Configurations}
\textbf{Datasets.} We validate the effectiveness of 4DGC using three real-world datasets: N3DV dataset \cite{li2022neural}, MeetRoom dataset \cite{streaming}, and  Google Immersive dataset \cite{broxton2020immersive}. Each dataset reserves 1 camera view for testing, with the remaining views used for training. 

\textbf{Implementation.} Our experimental setup includes an Intel(R) Xeon(R) W-2245 CPU @ 3.90GHz and an RTX 3090 graphics card. We set the resolution levels of the multi-resolution motion grid to $L=3$. During training, the initial settings are as follows: $\lambda_1$ is set to $0.0003, 0.0001,0.00005,0.00001$ to achieve different bitrates, while $\lambda_{2}$ is set to $0.2$. The number of iterations for motion estimation and Gaussian compensation is set to $400$ and $100$, respectively. 


\textbf{Metrics.} To evaluate the compression performance of our method on the experimental datasets, we use Peak Signal-to-Noise Ratio (PSNR) and Structural Similarity Index (SSIM) \cite{1284395} as quality metrics, along with bitrate measured in MB per frame. For comprehensive RD performance analysis, we apply Bjontegaard Delta Bit-Rate (BDBR) and Bjontegaard Delta PSNR (BD-PSNR) \cite{2007An}. Rendering efficiency is assessed by measuring the frames rendered per second (FPS).

\begin{table}[]
\centering
\setlength{\tabcolsep}{6.6pt} 
\renewcommand{\arraystretch}{0.95}
\caption{Quantitative comparison on the N3DV \cite{li2022neural} dataset. The PSNR, SSIM, size, and rendering speed are averaged over the whole 300 frames for each scene.}
\label{t1}
\vspace{-0.5em}
\scalebox{0.72}{
\begin{tabular}{c|cccc|c}
\hline
Method     & \begin{tabular}[c]{@{}c@{}}PSNR$\uparrow$\\ (dB)\end{tabular} & SSIM$\uparrow$  & \begin{tabular}[c]{@{}c@{}}Size$\downarrow$\\ (MB)\end{tabular} & \begin{tabular}[c]{@{}c@{}}Render$\uparrow$\\ (FPS)\end{tabular} & \begin{tabular}[c]{@{}c@{}}Streamable/\\ Variable-bitrate\end{tabular} \\ \hline
K-Planes \cite{kplanes}   & 29.91                                               & 0.920 & 1.0                                                    & 0.15                                                   & \ding{55}/\ding{55}                                                                     \\
                          HyperReel \cite{attal2023hyperreel}  & 31.10                                               & 0.938 & 1.2                                                    & 2.0                                                   & \ding{55}/\ding{55}                                                                     \\
                          MixVoxels \cite{10377294}  & 30.80                                               & 0.931 & 1.7                                                    & 16.7                                                   & \ding{55}/\ding{55}                                                                     \\
                          NeRFPlayer \cite{nerfplayer} & 30.69                                               & 0.931 & 17.7                                                   & 0.05                                                   & \ding{51}/\ding{55}                                                                   \\
                          StreamRF \cite{streaming}   & 30.61                                               & 0.930 & 7.6                                                    & 8.3                                                    & \ding{51}/\ding{55}             \\
                          ReRF \cite{rerf}   & 29.71                                               & 0.918 & 0.77                                                   & 2.0                                                    & \ding{51}/\ding{51}                                                         
                         \\
                          TeTriRF \cite{tetrirf}   & 30.65                                               & 0.931 & {\underline{0.76}}                                                    & 2.7                                                    & \ding{51}/\ding{51}                                                                    \\
                          D-3DG \cite{luiten2023dynamic}      & 30.67                                               & 0.931 & 9.2                                                    & \textbf{{460}}                                                    & \ding{51}/\ding{55}                                                                    \\
                          3DGStream \cite{sun20243dgstream}  & {\underline{31.54}}                                               & {\underline{0.942}} & 8.1                                                    & {\underline{215}}                                                    & \ding{51}/\ding{55}                                                                   \\
                          Ours       & \textbf{{31.58}}                                               & \textbf{{0.943}} & \textbf{{0.5}}                                                    & 168                                                    & \ding{51}/\ding{51}                                                                    \\ \hline
\end{tabular}
}
\end{table}
\begin{table}[]
\setlength{\tabcolsep}{2.4pt} 
\renewcommand{\arraystretch}{0.95}
\caption{Quantitative comparison on the MeetRoom dataset \cite{streaming} and Google Immersive  dataset \cite{broxton2020immersive}. 
}
\label{t2}
\vspace{-0.5em}
\scalebox{0.72}{
\begin{tabular}{c|c|cccc|c}
\hline
Dataset                                                                                & Method    & \begin{tabular}[c]{@{}c@{}}PSNR$\uparrow$\\ (dB)\end{tabular} & SSIM$\uparrow$  & \begin{tabular}[c]{@{}c@{}}Size$\downarrow$\\ (MB)\end{tabular} & \begin{tabular}[c]{@{}c@{}}Render$\uparrow$\\ (FPS)\end{tabular} & \begin{tabular}[c]{@{}c@{}}Streamable/\\ Variable-bitrate\end{tabular} \\ \hline
\multirow{5}{*}{\begin{tabular}[c]{@{}c@{}}MeetRoom\\ dataset \cite{streaming}\end{tabular}}            
                                                                                       & StreamRF \cite{streaming}  & 26.71                                               & 0.913 & 8.23                                                   & 10                                                     & \ding{51}/\ding{55}                                                          \\
                                                                                       & ReRF \cite{rerf}      & 26.43                                               & 0.911 & 0.63                                                  & 2.9                                                    & \ding{51}/\ding{51}                                                          \\
                                                                                       & TeTriRF \cite{tetrirf}  & 27.37                                                  & 0.917    & {\underline{0.61}}                                                     & 3.8                                                     & \ding{51}/\ding{51}                                                          \\
                                                                                       & 3DGStream \cite{sun20243dgstream} & \underline{{28.03}}                                               & \underline{{0.921}} & 8.21                                                   & \textbf{{288}}                                                    & \ding{51}/\ding{55}                                                          \\
                                                                                       & Ours      & \textbf{28.08}                                               & \textbf{{0.922}} & \textbf{{0.42}}                                                   & {\underline{213}}                                                    & \ding{51}/\ding{51}                                                         \\ \hline
\multirow{5}{*}{\begin{tabular}[c]{@{}c@{}}Google \\ Immersive\\ dataset \cite{broxton2020immersive}\end{tabular}} 

& StreamRF \cite{streaming}      & 28.14                                               & 0.929 & 10.24                                                  & 8.0                                                    & \ding{51}/\ding{55}                                                          \\
& ReRF \cite{rerf}      & 27.75                                               & 0.928 & 0.93                                                  & 1.4                                                    & \ding{51}/\ding{51}                                                          \\
& TeTriRF \cite{tetrirf}      & 28.53                                               & 0.931 & {\underline{0.83}}                                                  & 2.1                                                    & \ding{51}/\ding{51}                                                          \\                                                                                      
                                                                                       & 3DGStream \cite{sun20243dgstream} & \underline{{29.66}}                                                  & \textbf{{0.935}}    & 10.33                                                     & \textbf{{199}}                                                    & \ding{51}/\ding{55}                                                          \\
                                                                                       & Ours      & \textbf{29.71}                                                 & \textbf{0.935}    & \textbf{0.61}                                                     & \underline{145}                                                    & \ding{51}/\ding{51}                                                         \\ \hline
\end{tabular}
}
\vspace{-1.0em}
\end{table}


\subsection{Comparison}
\textbf{Quantitative comparisons.} To validate the effectiveness of our method, we compare it against several state-of-the-art approaches including  K-Planes \cite{kplanes}, HyperReel \cite{attal2023hyperreel}, MixVoxels \cite{10377294}, NeRFPlayer \cite{nerfplayer}, StreamRF \cite{streaming}, ReRF \cite{rerf}, TeTriRF \cite{tetrirf}, D-3DG \cite{luiten2023dynamic} and 3DGStream \cite{sun20243dgstream}. Tab. \ref{t1} shows the detailed quantitative results on the N3DV dataset. It can be seen that our method outperforms other methods and achieves the best reconstruction quality with the lowest bitrate. Specifically, 3DGStream \cite{sun20243dgstream} requires \textbf{8.1 MB} to achieve a comparable quality level to our 4DGC, which only needs \textbf{0.5 MB}. Although TeTriRF \cite{tetrirf} achieves a similar bitrate, its reconstruction quality is lower due to separate optimization of representation and compression. By contrast, our approach jointly optimizes the entire framework through a rate-distortion trade-off, which significantly enhances compression performance. To demonstrate the generality of our method, we conduct experiments on the MeetRoom and Google Immersive datasets, providing a quantitative comparison against StreamRF \cite{streaming}, ReRF \cite{rerf}, TeTriRF \cite{tetrirf},  and 3DGStream \cite{sun20243dgstream}, as illustrated in Tab. \ref{t2}. Our method still outperforms others in PSNR, SSIM, and bitrate. 

Fig. \ref{fig:ablation_rd} illustrates the RD curves  of our 4DGC compared to ReRF \cite{rerf}, TeTriRF \cite{tetrirf}, and 3DGStream \cite{sun20243dgstream} across various sequences from the three datasets.  The RD curves clearly show that our 4DGC achieves the best RD performance across a wide range of bitrates. Furthermore, we  calculate the BDBR relative to  ReRF \cite{rerf} and TeTriRF \cite{tetrirf}, as presented in Tab. \ref{table:BDBR1}. On the N3DV dataset, our 4DGC achieves an average BDBR reduction of \textbf{68.59\%} compared to TeTriRF \cite{tetrirf}.  Similar BDBR savings of \textbf{40.71\%} and \textbf{59.99\%} are observed on the MeetRoom and Google Immersive datasets, respectively. Against ReRF \cite{rerf}, our 4DGC also demonstrates significantly better RD performance.

Tab. \ref{time} compares the computational complexity of our 4DGC with the state-of-the-art dynamic scene compression methods, ReRF \cite{rerf} and TeTriRF \cite{tetrirf}. Our 4DGC significantly improves computational efficiency, with a training time of 0.83min versus 42.73min for ReRF and 1.04min for TeTriRF. In rendering, 4DGC requires only 0.006s, vastly outperforming ReRF (0.502s) and TeTriRF (0.375s). For encoding and decoding, 4DGC achieves times of 0.72s and 0.09s, respectively, surpassing both ReRF and TeTriRF. These results highlight 4DGC as a more efficient solution for FVV compression.

\textbf{Qualitative comparisons.} We present a qualitative comparison with ReRF \cite{rerf}, TeTriRF \cite{tetrirf}, and 3DGStream \cite{sun20243dgstream} on the \textit{coffee\_martini} sequence from the N3DV dataset and the \textit{trimming} sequence from the MeetRoom dataset, as shown in Fig. \ref{fig:Comparison}. Our approach achieves comparable reconstruction quality to 3DGStream \cite{sun20243dgstream} at a substantially lower bitrate, achieving a compression rate exceeding \textbf{16x}. Compared to ReRF \cite{rerf} and TeTriRF \cite{tetrirf}, our 4DGC more effectively preserves finer details such as the head, window, bottles, and books in \textit{coffee\_martini} and the face, hand, plant, and scissor in \textit{trimming}, which are lost in the reconstructions of these two methods. This demonstrates that our 4DGC captures dynamic scene elements accurately and maintains high-quality detail in intricate objects while achieving a highly compact model size. 
\begin{table}[]
\setlength{\tabcolsep}{12.5pt} 
\renewcommand{\arraystretch}{0.95}
\centering
\caption{The BDBR and BD-PSNR results of our 4DGC and ReRF \cite{rerf} when compared with TeTriRF \cite{tetrirf} on different datasets.}
\label{table:BDBR1}
\vspace{-0.5em}
\scalebox{0.75}{
\begin{tabular}{c|c|cc}
\hline
Dataset                   & Method & BDBR(\%)$\downarrow$   & BD-PSNR(dB)$\uparrow$ \\ \hline
\multirow{2}{*}{N3DV \cite{li2022neural}}     & ReRF    & 371.10 & -0.78   \\
                          & Ours   & \textbf{{-68.59}} & \textbf{{1.12}}    \\ \hline
\multirow{2}{*}{MeetRoom \cite{streaming}} & ReRF   & 134.69 & -0.99  \\
                          & Ours   & \textbf{{-40.71}} & \textbf{{0.55}}    \\ \hline
\multirow{2}{*}{\begin{tabular}[c]{@{}c@{}}Google\\ Immersive \cite{broxton2020immersive}\end{tabular}}   & ReRF   & 324.91 & -0.93   \\
                          & Ours   & \textbf{{-59.99}} & \textbf{{1.03}}    \\ \hline
\end{tabular}
}
\end{table}
\subsection{Ablation Studies}
We conduct three ablation studies to evaluate the effectiveness of motion estimation, Gaussian compensation, and joint optimization of representation and compression by disabling each component individually during training.  In the first study, we apply motion estimation but exclude Gaussian compensation. In the second, we omit motion estimation, training only the compensated Gaussians for each frame based on the previous frame. In the final study, we separately train the motion-aware representation and entropy models rather than optimizing them jointly.

The RD curves of the ablation studies are shown in Fig. \ref{fig:ablation_rd}. These curves illustrate that disabling motion estimation, Gaussian compensation or joint optimization results in reduced RD performance across various bitrates, underscoring the importance of these modules. Additionally, the minus BD-PSNR values observed in the three experiments compared to 4DGC, as shown in Tab. \ref{t3}, further confirm the effectiveness of our 4DGC in compressing dynamic scenes.

\begin{table}[]
\setlength{\tabcolsep}{9pt} 
\renewcommand{\arraystretch}{0.95}
\centering
\caption{Complexity comparison of our 4DGC method with dynamic scene compression methods, ReRF \cite{rerf} and TeTriRF \cite{tetrirf}.}
\label{time}
\vspace{-0.5em}
\scalebox{0.75}{
\begin{tabular}{c|cccc}
\hline
Method    & Train(min) & Render(s) & Encode(s) & Decode(s) \\ \hline
ReRF \cite{rerf}    & 42.73      & 0.502     & 3.03      & 0.28      \\
TeTriRF \cite{tetrirf} & 1.04       & 0.375     & 0.79      & 0.31      \\
4DGC    & \textbf{0.83}       & \textbf{0.006}     & \textbf{0.72}      & \textbf{0.09}      \\ \hline
\end{tabular}
}

\end{table}

\begin{figure}
\centering
\includegraphics[width=\linewidth]{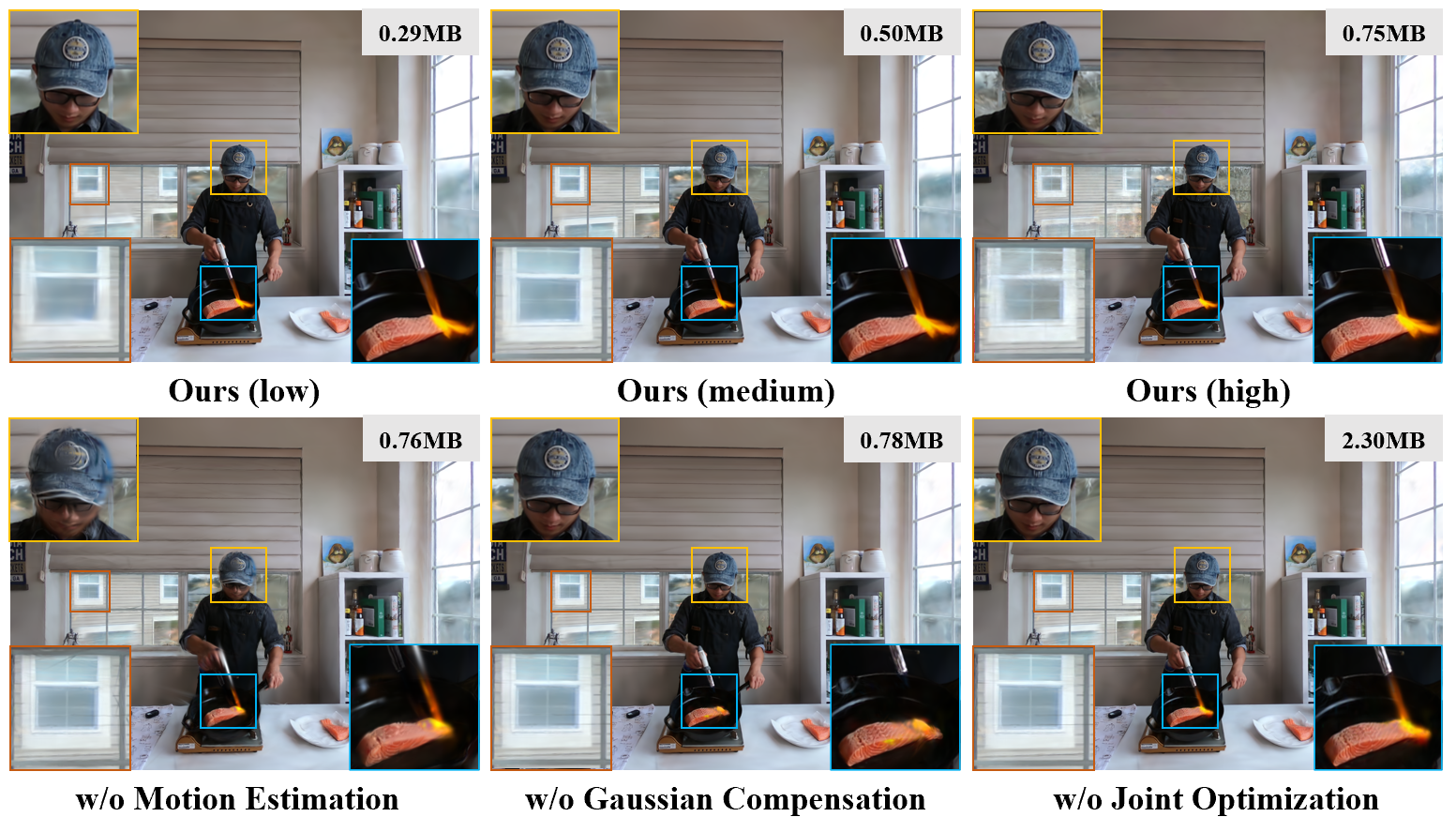}
\caption{Qualitative results of 4DGC and its variants. Excluding any module leads to lower reconstruction quality and increased bitrate.}
\label{fig:ablation_pic}  
\vspace{-1.em}
\end{figure}

Fig. \ref{fig:ablation_pic} illustrates a qualitative comparison of the complete 4DGC at different bitrates against its variants. When motion estimation is absent, details are added directly to the initial frame without tracking object motion, resulting in overlapping artifacts and increased temporal redundancy. The variant without Gaussian compensation struggles to capture newly appearing regions, such as fire eruptions. Moreover, The variant lacking joint optimization disregards the distribution characteristics of both the motion grid and compensated Gaussians, limiting the encoder's ability to achieve low entropy. These findings highlight the effectiveness of our motion estimation, Gaussian compensation, and joint optimization in the 4DGC.
\begin{table}[]
\setlength{\tabcolsep}{6.2pt} 
\renewcommand{\arraystretch}{0.95}
\centering
\caption{The BD-PSNR results of the ablation studies when compared with our full method on different datasets.}
\label{t3}
\vspace{-0.5em}
\scalebox{0.74}{
\begin{tabular}{c|ccc}
\hline
   & N3DV  & MeetRoom  & Google Immersive  \\ \hline
w/o Motion Estimation    & -1.86 dB & -1.13 dB    & -1.62 dB            \\
w/o Gaussian Compensation    & -0.23 dB & -0.22 dB    & -0.09 dB            \\
w/o Joint Optimization & -0.28 dB & -0.34 dB    & -0.40 dB            \\ \hline
\end{tabular}
}
\vspace{-1em}
\end{table}



Furthermore, we analyze the average bit consumption for keyframe and inter-frame  under different $\lambda_1$ configurations, as shown in Tab. \ref{size}. The significantly lower bit consumption for inter-frames demonstrates the effectiveness of our dynamic modeling in reducing inter-frame redundancy and lowering inter-frame bitrates.

\begin{table}[]
\setlength{\tabcolsep}{2pt} 
\renewcommand{\arraystretch}{0.95}
\centering
\caption{Analysis of average bit consumption for keyframe and inter-frame with different $ \lambda_1 $ on the N3DV dataset.}
\label{size}
\vspace{-0.5em}
\scalebox{0.75}{
\begin{tabular}{c|cccc}
\hline
                  & $\lambda_1 = 0.00001$ & $\lambda_1 = 0.00005$ & $\lambda_1 = 0.0001$ & $\lambda_1 = 0.0003$ \\ \hline
keyframe (MB)    & 17.3   & 14.4   & 10.6    & 7.3     \\
inter-frame (MB) & 1.21   & 0.78   & 0.48    & 0.23    \\ \hline
\end{tabular}
}
\vspace{-1em}
\end{table}

\section{Discussion}

\quad \textbf{Limitation.} As a novel and efficient rate-aware compression framework tailored for 4D Gaussian-based Free-Viewpoint Video, our method has several limitations. First, it relies on the reconstruction quality of the first frame, where poor initialization may degrade overall performance. Second, our method depends on multi-view video input and struggles with sparse-view reconstruction.Finally, the decoding speed is slower compared to rendering, which could be improved using advanced entropy decoding techniques.

\textbf{Conclusion.} We propose a novel rate-aware compression framework tailored for 4D Gaussian-based Free-Viewpoint Video. Leveraging a motion-aware 4D Gaussian representation, 4DGC effectively captures inter-frame dynamics and spatial details while sequentially reducing temporal redundancy. Our end-to-end compression scheme incorporates an implicit entropy model combined with rate-distortion tradeoff parameters, enabling variable bitrates while jointly optimizing both representation and entropy model for enhanced performance. Experiments show that 4DGC not only achieves superior rate-distortion performance but also adapts to variable bitrates, supporting photorealistic FVV applications with reduced storage and bandwidth requirements in AR/VR contexts.


\section{Acknowledgements}

This work was supported by National Natural Science Foundation of China (62271308), STCSM (24ZR1432000, 24511106902, 24511106900, 22511105700, 22DZ2229005), 111 plan (BP0719010) and State Key Laboratory of UHD Video and Audio Production and Presentation.\par
{
    \small
    \bibliographystyle{ieeenat_fullname}
    \bibliography{main}
}

\clearpage
\setcounter{page}{1}
\maketitlesupplementary

\section{More Implementation Details}
\label{sec:rationale}
\subsection{Entropy Model} Here, we provide a detailed introduction to our implicit entropy model. This model estimates entropy by learning a cumulative distribution function (CDF) that represents the probability distribution of the input data. The model comprises multiple layers, each parameterized by weight matrices, biases, and scaling factors. These parameters transform the input tensor through a series of operations, such as matrix multiplication and bias addition, combined with non-linear activations like softplus. Each layer progressively refines the input values to approximate a CDF, capturing the cumulative probability distribution of the data.

To maintain precision, we avoid assuming any predefined distribution for the data. Instead, we construct a novel distribution within the entropy model to closely approximate the actual data distribution. Specifically, the entropy model computes cumulative logits for values just below and above the actual data point. This approach enables the model to capture the probability interval that contains the data point, thereby improving estimation accuracy. The likelihood, which represents the probability of the input within this interval, is calculated as the difference between the sigmoid activations of these cumulative logits. This likelihood is then incorporated into the overall loss function during training.

When the training process is complete, we apply quantization and a range coder for entropy coding to further compress the data volume and generate the bitstream. The entropy model itself occupies about 100 KB per frame, which is relatively large compared to the compressed motion grid $\mathbf{M}_t$ and the compensated Gaussians $\Delta \mathbf{G}_t$. Therefore, we analyze the actual distribution of the data prior $\omega_t$ before entropy coding and transmit $\omega_t$ instead of the entire entropy model. This approach further reduces the size of each frame.

The process of entropy encoding can be represented as follows:
\begin{equation}
\begin{aligned}
\mathbf{Q}(x) &= \left\lfloor q \cdot x + 0.5 \right\rfloor,\\
B_t &= \mathbf{E} \left(\mathbf{Q}\left(x\right)-\mathbf{Q}\left( \min(x)\right); \omega_t \right).
\end{aligned} \label{eq:ME}
\end{equation}
Here, $x$ represents the data to be compressed, which includes $\mathbf{M}_t $ and $ \Delta \mathbf{G}_t$. The bitstream after entropy encoding, denoted as $B_t$, consists of $B_t^\mathbf{M} $ and $ B_t^{\Delta\mathbf{G}} $, corresponding to $\mathbf{M}_t $ and $\Delta \mathbf{G}_t $, respectively. The range encoder is represented by $\mathbf{E}$.

To enable compression into the int8 format, we convert the compressed data into non-negative values and subsequently restore it to its original range during decompression. The variable $q$ denotes the quantization parameter. During quantization, the data is multiplied by $q$, which effectively expands its range and subtly enhances the precision of the quantization process. On the decoding side, the entropy decoding process can be expressed as follows:

\begin{equation}
\begin{aligned}
\hat{x}=\frac{\mathbf{D}\left(B_t; \omega_t \right)+\mathbf{Q}\left(\min(x)\right)}{q},
\end{aligned} \label{eq:ME}
\end{equation}

\begin{figure*}[ht]
\centering
\includegraphics[width=\linewidth]{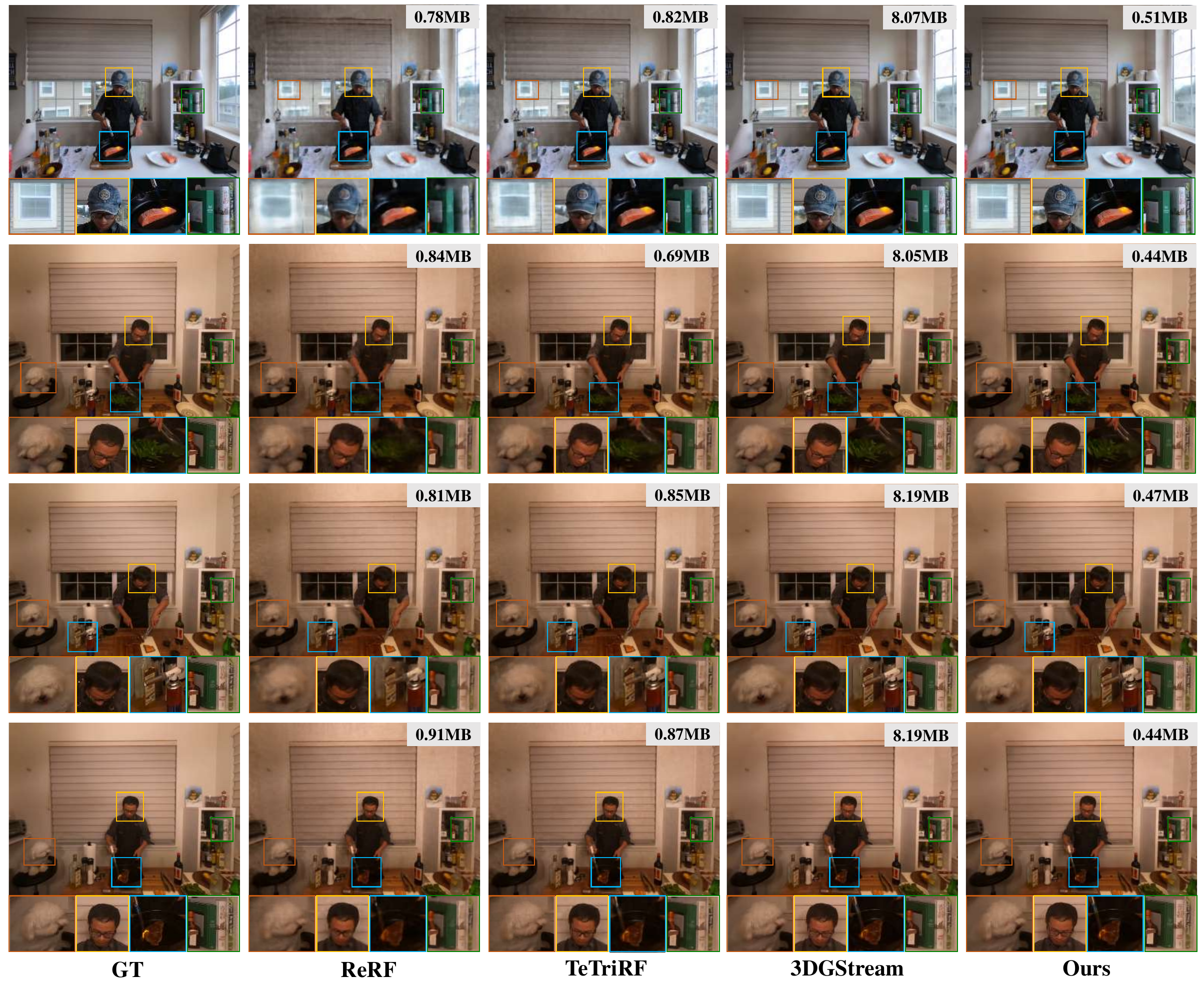}
\caption{More Qualitative comparison on more sequences of the N3DV dataset against ReRF, TeTriRF, and 3DGStream.}
\label{fig:supComparison}

\end{figure*}

where $\mathbf{D}$ is the range decoder. Therefore, for each inter-frame, the data to be transmitted includes the bitstream \( B_t^\mathbf{M} \) and \( B_t^{\Delta \mathbf{G}} \), and the data distributions $\omega_t = \{\omega_t^\mathbf{M},\omega_t^{\Delta \mathbf{G}}\} $. The size of each of these components is detailed in Tab. \ref{sup_size}.

\begin{table}[]
\centering
\caption{The average size of each component of inter-frames on the N3DV dataset.}
\label{sup_size}
\begin{tabular}{c|ccccc}
\hline
            & $B_t^\mathbf{M}$ & $B_t^{\Delta \mathbf{G}}$ & $\omega_t^\mathbf{M}$ & $\omega_t^{\Delta \mathbf{G}}$ & Total  \\ \hline
Size (KB) & 164.72            & 65.88                     & 0.25                  & 0.17                           & 231.02 \\ \hline
\end{tabular}
\vspace{-1em}
\end{table}

\subsection{Hyperparameters Settings}
In this section, we provide a more detailed explanation of the hyperparameter settings for the two aspects.

\textbf{Model Parameter Settings}: We use two shared global lightweight MLPs $\Phi_{\boldsymbol{\mu}}$ and $\Phi_{\mathbf{R}}$, both with an input dimension of 20, a hidden layer size of 64, and output dimensions of 3 and 4, respectively. For the multi-resolution motion grid \( \mathbf{M}_t \), we use feature grid channels of 4, 4, and 2, with dimensions \( 32^3 \), \( 64^3 \), and \( 128^3 \).

\textbf{Gaussian Compensation Parameter Settings}: For Gaussian compensation, we choose \( \tau_{g} = 0.0001 \), \( \tau_{\mu} = 0.08 \), and \( \tau_{R} = \frac{\pi}{4} \). We only apply the second type of Gaussian compensation to the Gaussians whose scale \( \mathbf{s} > -0.01 \). (The actual scale is activated using the exponential function.) To prevent an excessive increase in the number of Gaussians, we filter out Gaussians with opacity \( \alpha < 0.01 \) after stage 2.

\begin{figure*}[ht]
\centering
\includegraphics[width=\linewidth]{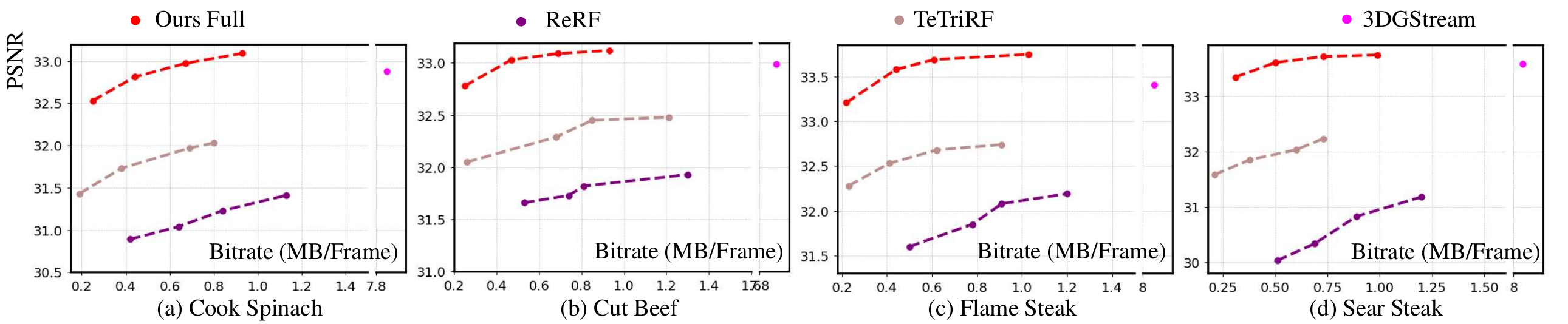}
\caption{More Rate-distortion curves across different sequences of N3DV datasets. }
\label{sup_fig1}
\end{figure*}
\section{More Results}
\subsection{Quantitative Results}
We provide a quantitative comparison of image quality, measured by PSNR, and model size across all scenes in the N3DV dataset in Tab. \ref{tab:sup_t1}. To further demonstrate the variable bitrate characteristic and superior RD performance of our method, we present additional RD curves for more sequences in Fig. \ref{sup_fig1}.
\begin{table*}[ht]
\centering
\caption{Quantitative comparison of average PSNR values(dB) and model size(MB) across all sequences in the N3DV dataset.}
\label{tab:sup_t1}
\begin{tabular}{c|cccccc|c}
\hline
Method    & \begin{tabular}[c]{@{}c@{}}Coffee\\ Martini\end{tabular} & \begin{tabular}[c]{@{}c@{}}Cook\\ Spinach\end{tabular} & \begin{tabular}[c]{@{}c@{}}Cut\\ Beef\end{tabular} & \begin{tabular}[c]{@{}c@{}}Flame\\ Salmon\end{tabular} & \begin{tabular}[c]{@{}c@{}}Flame\\ Steak\end{tabular} & \begin{tabular}[c]{@{}c@{}}Sear\\ Steak\end{tabular} & Mean       \\ \hline
StreamRF  & 27.77/9.34                                               & 31.54/7.48                                             & 31.74/7.17                                         & 28.19/7.93                                             & 32.18/7.02                                            & 32.29/6.88                                           & 30.61/7.64 \\
ReRF      & 26.24/0.79                                               & 31.23/0.84                                             & 31.82/0.81                                         & 26.80/0.78                                             & 32.08/0.91                                            & 30.03/0.51                                           & 29.71/0.77 \\
TeTriRF   & 27.10/0.73                                               & 31.97/0.69                                             & 32.45/0.85                                         & 27.61/0.82                                             & 32.74/0.87                                            & 32.03/0.60                                           & 30.65/0.76 \\
3DGStream & 27.96/8.00                                               & \textbf{32.88}/8.05                                             & 32.99/8.19                                         & \textbf{28.52}/8.07                                             & 33.41/8.19                                            & 33.58/8.16                                           & 31.54/8.11 \\
Ours      & \textbf{27.98}/\textbf{0.58}                                               & 32.81/\textbf{0.44}                                              & \textbf{33.03}/\textbf{0.47}                                         & 28.49/\textbf{0.51}                                             & \textbf{33.58}/\textbf{0.44}                                            & \textbf{33.60}/\textbf{0.50}                                           & \textbf{31.58}/\textbf{0.49} \\ \hline
\end{tabular}
\end{table*}
\subsection{Qualitative Results}
We have prepared more qualitative comparison results in the Fig. \ref{fig:supComparison}. 

\end{document}